# Anytime Decision Making with Imprecise Probabilities

**Michael Pittarelli**
Computer Science Department
SUNY Institute of Technology
Utica, NY 13504-3050
mike@sunyit.edu

## Abstract

This paper examines methods of decision making that are able to accommodate limitations on both the form in which uncertainty pertaining to a decision problem can be realistically represented and the amount of computing time available before a decision must be made. The methods are *anytime algorithms* in the sense of Boddy and Dean [1989]. Techniques are presented for use with Frisch and Haddawy's [1992] anytime deduction system, with an anytime adaptation of Nilsson's [1986] probabilistic logic, and with a probabilistic database model.

## 1 ANYTIME ALGORITHMS FOR DECISION MAKING

In general, an anytime algorithm [Boddy and Dean, 1989] provides output at each step of its execution; the output improves, in some sense, the longer the algorithm runs. For a decision problem under risk, where the probability of each of the conditions is known and is represented as a single real number, an anytime decision algorithm might consider the actions in some sequence and output at each step the action with highest expected utility among those so far examined.

A more realistic representation of uncertainty is by means of intervals of real numbers [Kyburg, 1992]. A reasonable criterion might interpret the intervals as linear inequality constraints determining a set of real-valued probability functions over the conditions and identify an action as inadmissible if there does not exist an element of the set relative to which the action maximizes expected utility [Levi, 1980]. This may be determined for each action by checking for the existence of a feasible solution to a linear program [Pittarelli, 1991]. Any action not yet ruled out as inadmissible by some time may be classified (pos-sibly incorrectly) as admissible relative to the current collection of probability intervals. At a higher level, the set of intervals may be iteratively refined (e.g., for confidence intervals, by decreasing the confidence level [Loui, 1986]) and the process of testing for inadmissibility repeated for each refinement, for the elements of the most recently computed admissible set.

Anytime methods for probabilistic inference have recently been developed [Frisch and Haddawy, 1992]. These determine an interval of probability for a conclusion from a collection of premises each of which has an associated probability interval. The interval of probability for the conclusion is initially [0,1] and is narrowed with each application of a rule of inference. This approach differs from [Nilsson, 1986] which requires the construction of a potentially very large linear program for determining the endpoints of the probability interval for the entailed sentence. However, it is straightforward to base an anytime inference technique directly on Nilsson's methods, by considering only subsets of the given set of premises. We will discuss methods of anytime decision making that utilize both forms of probabilistic inference.

Probabilistic databases [Pittarelli, 1994], that is, collections of contingency tables of joint frequencies or probabilities for finite variables, may also provide linear constraints relevant to a decision problem. There is again a tradeoff between the tightness of the constraint that can be inferred and the cost of doing the inference. We will discuss methods of anytime decision making utilizing probabilistic databases also.

## 2 ANYTIME DEDUCTION AND DECISIONS

Frisch and Haddawy [1992] have developed a system of deduction for probabilistic logic based on inference rules. The rules may be employed to compute increasingly narrow probability intervals for the conclusion of an argument. At any time, the currently computed interval is correct in the sense that it contains the narrowest interval computable from the probabilities associated with all of the premises. Thus, there is a tradeoff between the precision of an entailed probability interval and the time required to compute it. This feature makes anytime deduction especially suitable for use by "resource-bounded" systems [Horvitz et al, 1989]; as Frisch and Haddawy point out, however, how to control the time/precision tradeoff depends on the particular decision situation in which the system finds itself.

Frisch and Haddawy's anytime deduction, Nilsson's probabilistic logic (unless maximum entropy or related techniques are used), and related systems, produce probability intervals for entailed sentences. A criterion applicable to decision problems in which probabilities are given as intervals and that reduces to standard maximization of expected utility when the intervals reduce to point values is Levi's *E-admissibility* criterion [Levi, 1980]: all and only those actions maximizing expected utility relative to some member of the set of (point valued) probability functions compatible with (or actually representing [Kyburg, 1992]) current beliefs are admissible.

We will consider decision problems consisting of a set of actions $A = \{a_1, \ldots, a_m\}$ and a set of mutually exclusive and exhaustive conditions $C = \{c_1, \ldots, c_n\}$ such that

$$p(c_j \mid a_i) = p(c_j), \text{ for all } i, j.$$

(Jeffrey [1976] shows how a problem in which conditions are not probabilistically independent of actions can be converted to an equivalent problem in which they are.)

A family $D$ of subsets of the $(n-1)$-dimensional simplex $P_C$ of all probability distributions over $C$ may be defined as

$$D = \{D(a_1), \ldots, D(a_m)\},$$

where

$$D(a_i) = \{p \in P_C \mid \sum_{j=1}^{n} p(c_j) \times U(a_i, c_j) \geq \sum_{j=1}^{n} p(c_j) \times U(a_k, c_j), k = 1, \ldots, m\}.$$

$D(a_i)$ is the set of probability functions relative to each of which action $a_i$ maximizes expected utility, and is referred to as the *domain* of $a_i$ [Starr, 1966].

Each domain is convex; domains may intersect at faces; and $\bigcup_{a \in A} D(a) = P_C$.

Suppose the current belief state regarding probabilities of conditions is represented as $K \subseteq P_C$. An action $a$ is E-admissible if and only if $K \cap D(a) \neq \emptyset$. The fewer admissible actions, the better. The number of admissible actions decreases monotonically with the size of $K$:

$$K_1 \subseteq K_2 \to K_1 \cap D(a_i) \subseteq K_2 \cap D(a_i).$$

Thus, there will be a tradeoff between the quality of a decision — i.e., the number of admissible actions among which choice must be made using non-probabilistic criteria — and the computational expense of shrinking $K$. Reduction in the size of $K$ may be achieved by applying additional inference rules; by adding premises, thereby enlarging the linear constraint system; by "extending" and "projecting" larger and larger portions of a probabilistic database; etc. (Of course, the *perceived* quality of a decision may be enhanced without this much work: invoke the maximum entropy principle. This approach will be criticized below.) Computing aimed at reducing the number of admissible actions may be interleaved with analysis of the type proposed by Horvitz et al. [1989] to determine whether the cost of such computation exceeds its expected value.

Consider a decision the outcome of which is contingent on the truth or falsity of a single probabilistically entailed sentence: "It will rain this afternoon". Suppose the actions under consideration are "Go to the beach" and "Do not go to the beach". Utilities of the four possible outcomes are:

$$U(\text{Go, Rain}) = 0,$$
$$U(\text{Go, No rain}) = 1,$$
$$U(\text{Do not, Rain}) = 0.8,$$
$$U(\text{Do not, No rain}) = 0.2.$$

The agent's knowledge is represented in part by the propositions and associated probability intervals:

(1) $p(\text{Temperature} > 85) \in [.95,1]$
(2) $p(\text{Temperature} > 85 \to \text{Rain}) \in [.4,.6]$
(3) $p((\text{B. pressure} < 30 \text{ \& Humidity} > 80) \to \text{Rain}) \in [.65,.95]$
(4) $p(\text{B. pressure} < 30) \in [.95,1]$
(5) $p(\text{Humidity} > 80) \in [.95,1]$
(6) $p(\text{August} \to \text{Rain}) \in [.2,1]$
(7) $p(\text{August}) \in [1,1]$.

Both "Go to the beach" and "Do not go to the beach" have non-empty domains: "Go" maximizes expected utility when $p(\text{Rain}) \leq 0.5$; "Do not" does for $p(\text{Rain}) \geq 0.5$. Neither can be ruled out *a priori*. However, Frisch and Haddawy's probabilistic infer-ence rules may be applied one-at-a-time to narrow the interval for $p(\text{Rain})$ until a single admissible action emerges, or it is no longer economical to continue refining (e.g., the last train to the beach is about to leave) and a choice among the admissible actions must be made using some other criterion (e.g., choose at random [Elster, 1989], use maximin, maximize expected utility relative to the midpoint of the probability interval, etc.).

Initially, we can deduce

(8) $p(\text{Rain}) \in [0,1]$,

from the "Trivial derivation" rule:

$$\vdash p(\alpha \mid \delta) \in [0,1].$$

We may next apply "Forward implication propagation",

$$p(\beta \mid \delta) \in [x,y], \; p(\beta{\rightarrow}\alpha \mid \delta) \in [u,v] \vdash$$
$$p(\alpha \mid \delta) \in [min(0, x+u-1), v],$$

to statements (1) and (2), yielding

(9) $p(\text{Rain}) \in [.35, .6]$.

Although it does not have any effect at this stage, the "Multiple derivation" rule should be applied to maintain the tightest interval for the "target" sentence:

$$p(\alpha \mid \delta) \in [x,y], \; p(\alpha \mid \delta) \in [u,v] \vdash$$
$$p(\alpha \mid \delta) \in [max(x,u), min(y,v)].$$

Since $.5 \in [.35, .6]$, both actions remain admissible. Next, "Conjunction introduction",

$$p(\alpha \mid \delta) \in [x,y], \; p(\beta \mid \delta) \in [u,v]$$
$$\vdash p(\alpha \; \& \; \beta \mid \delta) \in [max(0, x+u-1), min(y,v)]$$

is applied to statements (4) and (5), yielding

(10) $p(\text{B. pressure} < 30 \; \& \; \text{Humidity} > 80)$
$$\in [.9, 1].$$

Applying forward implication propagation to statements (3) and (10) gives

(11) $p(\text{Rain}) \in [.55, .95]$

Although combining statement (11) with statement (9) via the multiple derivation rule will further narrow the target interval, there is no need to do so; nor is there any need to consider statements (6) and (7). "Do not go" has emerged as uniquely admissible:

$$D(\text{Do not go}) = \{p{\in}P_{\{Rain, \, No \, rain\}} \mid p(\text{Rain}) \geq .5\},$$
$$D(\text{Go}) = \{p{\in}P_{\{Rain, \, No \, rain\}} \mid p(\text{Rain}) \leq .5\},$$
$$D(\text{Go}){\cap}\{p{\in}P_{\{Rain, \, No \, rain\}} \mid p(\text{Rain}){\in}[.55, .95]\} = \varnothing.$$
$$D(\text{Do not go}) \cap$$
$$\{p{\in}P_{\{Rain, \, No \, rain\}} \mid p(\text{Rain}){\in}[.55, .95]\} \neq \varnothing.$$

## 3 NILSSON'S PROBABILISTIC LOGIC AND DECISION MAKING

Nilsson's methods may be modified to yield an anytime procedure for decision making. Rather than construct the linear system corresponding to the full set of sentences, increasingly larger systems may be constructed by adding sentences to the subset currently in use until a uniquely admissible action emerges or it is necessary to choose among the currently admissible actions.

This may be illustrated with the sentences and decision problem above. Suppose sentences (3) and (5) are chosen for the first iteration. Using Nilsson's "semantic tree" method, five sets of possible worlds are identified. Both actions are E-admissible. "Go" is E-admissible because there exist feasible solutions to the system of linear inequalities below, where $p_i$ is the probability of set $w_i$ of possible worlds; "Rain" is true in sets $w_3$ and $w_5$, "Humidity > 80" is true in sets $w_1$, $w_4$ and $w_5$, etc.:

$$p_1 + p_2 + p_3 + p_4 + p_5 = 1$$
$$p_2 + p_3 + p_4 + p_5 \geq 0.65$$
$$p_2 + p_3 + p_4 + p_5 \leq 0.95$$
$$p_1 + p_4 + p_5 \geq 0.95$$
$$(p_3 + p_5){\times}0 + (p_1 + p_2 + p_4){\times}1 \geq$$
$$(p_3 + p_5){\times}0.8 + (p_1 + p_2 + p_4){\times}0.2$$

"Do not go" is also E-admissible, since the system resulting from reversing the direction of the final inequality also has feasible solutions.

Now add sentence (4). The resulting 8 sets of possible worlds may be determined by expanding only the "live" terminal nodes of the semantic tree constructed at the first iteration. (To eliminate the need for a row interchange, the root of the initial tree should represent the target sentence. One may proceed in this way until it is no longer necessary to continue, possible to continue, or worth continuing. If the number of sets of worlds generated becomes excessive, Snow's compression method [1991] may be attempted.) "Do not go" is now identified as uniquely E-admissible; there exist feasible solutions to the system below, but not to the corresponding system for "Go":

$$p_1 + p_2 + p_3 + p_4 + p_5 + p_6 + p_7 + p_8 = 1$$
$$p_1 + p_2 + p_3 + p_4 + p_5 + p_6 + p_7 \geq 0.65$$
$$p_1 + p_2 + p_3 + p_4 + p_5 + p_6 + p_7 \leq 0.95$$
$$p_1 + p_2 + p_6 + p_8 \geq 0.95$$
$$p_1 + p_3 + p_5 + p_8 \geq 0.95$$
$$(p_1 + p_2 + p_3 + p_4){\times}0.8 + (p_5 + p_6 + p_7 + p_8){\times}0.2$$
$$\geq (p_1 + p_2 + p_3 + p_4){\times}0 + (p_5 + p_6 + p_7 + p_8){\times}1$$

## 4 DECISIONS WITH MULTIPLE CONDITIONS

Frisch and Haddawy's system is applicable to decision problems with an arbitrary number $n$ of mutually exclusive conditions. The $(\frac{1}{2}n(n-1)+1)$ statements

$p(c_1 v \cdots v c_n) \in [1,1]$
$p(c_1 \& c_2) \in [0,0]$
.
.
.
$p(c_{n-1} \& c_n) \in [0,0]$

must be included. Intervals must be maintained for each of the conditions $c_j$. The soundness of Frisch and Haddawy's inference rules guarantees that, at any time, the interval $[l_j, u_j]$ associated with any $c_j$ is a superset of the tightest interval entailed (algebraically) by the full collection of sentences. Thus, the sharpest intervals available at any time yield a linear system from which it can be determined whether an action would not be E-admissible relative to the sharper probability bounds computable at any later time; action $a_i$ is (ultimately) admissible only if there exist feasible solutions to

$p(c_1) + \cdots + p(c_n) = 1$
$p(c_1) \geq l_1$
$p(c_1) \leq u_1$
.
.
.
$p(c_n) \geq l_n$
$p(c_n) \leq u_n$
$p(c_1) \times U(a_i, c_1) + \cdots + p(c_n) \times U(a_i, c_n) \geq$
$\quad p(c_1) \times U(a_1, c_1) + \cdots + p(c_n) \times U(a_1, c_n)$
.
.
.
$p(c_1) \times U(a_i, c_1) + \cdots + p(c_n) \times U(a_i, c_n) \geq$
$\quad p(c_1) \times U(a_m, c_1) + \cdots + p(c_n) \times U(a_m, c_n),$

where $l_j$ and $u_j$ are the current bounds on $p(c_j)$.

Information may be lost if probability intervals are computed separately for each of the conditions in a decision problem with more than two conditions. There are convex polytopes of probability distributions over $n > 2$ conditions such that the solution set of the linear system obtained by combining the unicity constraint with the inequalities corresponding to the tightest probability bounds inferable from the polytope for the conditions is a proper superset of it. The intersection of the domain of an action with the original polytope may be empty, although its intersection with the solution set is not. Thus, actions that are not E-admissible may not be identified as such, resulting in unnecessary indeterminateness.

Nilsson's semantic tree method can be adapted to take into account the mutual exclusivity and exhaustiveness of multiple (i.e., more than two) conditions in a decision problem. The first $n$ levels of the tree will correspond to the $n$ conditions. (This facilitates the anytime adaptation of Nilsson's methods discussed above.) At level $n$ there will be $n$ live nodes, one for each of the assignments in which exactly one of the conditions is true. The remaining levels of the tree are constructed as usual.

For example, with conditions $c_1$, $c_2$ and $c_3$, an arbitrary number $m \geq 2$ of actions $a_i$, and data $p(B \rightarrow c_1) \in [0.9, 1]$ and $p(B) \in [0.8, 1]$, there are 6 sets of possible worlds, corresponding to the matrix

$$\begin{array}{ll}
1\,1\,0\,0\,0\,0 & (c_1) \\
0\,0\,1\,1\,0\,0 & (c_2) \\
0\,0\,0\,0\,1\,1 & (c_3) \\
1\,1\,1\,0\,1\,0 & (B \rightarrow c_1) \\
1\,0\,0\,1\,0\,1 & (B)
\end{array}$$

Action $a_i$ is E-admissible iff there exist feasible solutions to the system of linear inequalities:

$p_1 + \cdots + p_6 = 1$
$p_1 + p_2 + p_3 + p_5 \geq 0.9$
$p_1 + p_4 + p_6 \geq 0.8$
$(p_1 + p_2) \times U(a_i, c_1) + (p_3 + p_4) \times U(a_i, c_2)$
$\quad + (p_5 + p_6) \times U(a_i, c_3) \geq$
$\quad\quad (p_1 + p_2) \times U(a_1, c_1) + (p_3 + p_4) \times U(a_1, c_2)$
$\quad\quad\quad + (p_5 + p_6) \times U(a_1, c_3)$
.
.
.
$(p_1 + p_2) \times U(a_i, c_1) + (p_3 + p_4) \times U(a_i, c_2)$
$\quad + (p_5 + p_6) \times U(a_i, c_3) \geq$
$\quad\quad (p_1 + p_2) \times U(a_m, c_1) + (p_3 + p_4) \times U(a_m, c_2)$
$\quad\quad\quad + (p_5 + p_6) \times U(a_m, c_3)$

## 5 MAXIMUM ENTROPY AND PROBABILISTIC LOGIC

Nilsson [1986] shows how to maximize entropy within the set of probability distributions over the possible worlds in order to compute a point-valued probability for an entailed sentence. The maximum entropy estimate of the probability of the entailed sentence is the sum of the components of the maximum entropy distribution corresponding to the worlds in which the sentence is true. Point-valued probabilities for each of the conditions in a decision problem are computable also from the distribution over the possible worlds maximizing entropy.

If an action maximizes expected utility relative to the maximum entropy estimate, it is guaranteed to be E-admissible relative to any set of distributions to which the estimate belongs. But, of course, the converse does not hold. E-admissible actions that, depending on one's philosophy of decision making, perhaps should be retained for further consideration, are eliminated. It may be that one of these actions uniquely maximizes expected utility relative to the (*pace, inter alios*, DeFinetti) true but unknown distribution.

If the maximum entropy distribution tends to be close, on some metric, to the actual distribution over the worlds, then its projection will tend to be close to the actual probability distribution over the conditions. (But note that the result of marginalizing the maximum entropy element of a set $K$ is not always the maximum entropy element of the set of marginals of elements of $K$.) The closer the estimate of the probabilities of the conditions is to the true distribution, the likelier it is that it will belong to one of the domains containing the true distribution. Thus, the likelier it is that an action maximizing utility relative to the true distribution will be selected.

How close can one expect the maximum entropy estimate to be to the true distribution over the possible worlds? If you accept Jaynes' *concentration theorem* [Jaynes, 1982], i.e., if probabilities are observed relative frequencies and sequences of observations are equiprobable *a priori* and the number of observations is infinite and if you accept difference in entropy as a distance measure, then the answer is "very". If you want to stick with metric distances and probabilities are allowed to be subjective, then it might be reasonable to ask how close the maximum entropy element is to the *centroid* of the set, which minimizes expected sum-of-squares error [MacQueen and Marschak, 1975], but is more expensive to calculate [Piepel, 1983].

When the set of distributions is either a singleton or the full probability simplex, the maximum entropy element is guaranteed to coincide with the centroid. It always coincides with the centroid for the *modus ponens* inference pattern: Let $x = p(P)$ and $y = p(P \rightarrow Q)$. There are four sets of possible worlds, the probabilities of which are the solutions to the system of equations:

$p_1 + p_2 + p_3 + p_4 = 1$
$p_1 + p_2 = x$
$p_1 + p_3 + p_4 = y.$

The solution set is either a single point (when $p(P){=}1$) or a line segment in $[0,1]^4$ with vertices

$(p_1, p_2, p_3, p_4) = (y-(1-x), (1-y), 1-x, 0)$
$(p_1, p_2, p_3, p_4) = (y-(1-x), (1-y), 0, 1-x).$

The centroid of the line segment is the average of the vertices, which coincides with the maximum entropy distribution calculated by Nilsson.

This will not always be the case. It does not even appear that one can expect the maximum entropy estimate to be especially close to the centroid (which one cares about if one wishes to minimize expected squared error). Consider the *conjunction* pattern of inference: from $A$ and $B$, infer $A$ & $B$. There are four sets of possible worlds: those in which $A$&$B$ is true, those in which $A$&$\overline{B}$ is true, etc. The set of solutions to the system is again either a single point or a line segment in $[0,1]^4$. Let $l(A\&B)$ and $u(A\&B)$ denote, respectively, the greatest lower and least upper bounds on $p(A\&B)$. Ordering components as

$(p(A\&B), p(A\&\overline{B}), p(\overline{A}\&B), p(\overline{A}\&\overline{B})),$

the solution set has vertices

$v^1 = (l(A\&B), p(A)-l(A\&B), p(B)-l(A\&B),$
$\quad (1-(p(A)+p(B)-l(A\&B)))$
$v^2 = (u(A\&B), p(A)-u(A\&B), p(B)-u(A\&B),$
$\quad (1-(p(A)+p(B)-u(A\&B))).$

The centroid is the average of the two vertices:

$$ce = (v^1 + v^2)/2.$$

The maximum entropy element coincides with the distribution computed under the assumption of probabilistic independence of $A$ and $B$:

$$m = (p(A) \times p(B), p(A) \times p(\overline{B}), p(\overline{A}) \times p(B),$$
$$p(\overline{A}) \times p(\overline{B})).$$

The *eccentricity* of an element of any non-unit solution set $K$ is the ratio between its (Euclidean) distance from the centroid and the maximum distance of any element of the set from the centroid:

$$ecc(p, K) = d(p, ce)/\max_{p \in K} d(p, ce).$$

The eccentricity will have a minimum value of 0 (when $p = ce$) and a maximum value of 1 (when $p$ is a vertex).

For conjunction entailment, it is possible for the value of $ecc(m, K)$ to be quite high. For example, when $p(A){=}0.9$ and $p(B){=}0.1$, $ecc(m, K){=}0.8$. The expected value of $ecc(p, K)$ for a randomly selected element $p$ of $K$ is $1/2$. Letting $<p(A), p(B)>$ range with uniform probability over $(0,1)^2$, the expected value of $ecc(m, K)$ is $1/3$. So, for conjunction entailment anyway, one cannot expect the maximum entropy approximation to be especially low-risk.

Kane [1990, 1991] has developed a method of computing the maximum entropy solution that is faster than that proposed by Nilsson. Deutsch-McLeish [1990] has determined conditions under which Nilsson's *projection approximation* (which is not, in general, the centroid of the solution set) coincides with the maximum entropy solution. These can be tested to determine whether the (much cheaper) projection approximation method can be substituted for direct maximization of entropy. But, as argued above, computing any type of point-valued estimate of condition probabilities for a decision problem is neither necessary nor wise.

## 6 ANYTIME DECISION MAKING WITH PROBABILISTIC DATABASES

A *probabilistic database* [Cavallo and Pittarelli, 1987; Pittarelli, 1994; Barbara *et al*, 1993] generalizes a relational database by replacing the characteristic function of a relation with a probability distribution (or probability intervals). For example, the tables below represent estimates of probabilities for various (joint) events on a typical August day in a fictitious developing country:

| Rain | No Phones | $p^1$ |
|------|-----------|-------|
| yes | true | 0.4 |
| yes | false | 0.1 |
| no | true | 0.2 |
| no | false | 0.3 |

| No Phones | Trains | $p^2$ |
|-----------|--------|-------|
| true | yes | 0.25 |
| true | no | 0.35 |
| false | yes | 0.25 |
| false | no | 0.15 |

| No Phones | Temperature | $p^3$ |
|-----------|-------------|-------|
| true | high | 0.45 |
| true | med | 0.1 |
| true | low | 0.05 |
| false | high | 0.25 |
| false | med | 0.1 |
| false | low | 0.05 |

| Temperature | Humidity | $p^4$ |
|-------------|----------|-------|
| high | high | 0.6 |
| high | low | 0.1 |
| med | high | 0.15 |
| med | low | 0.05 |
| low | high | 0 |
| low | low | 0.1 |

Suppose, again, that it must be decided whether or not to go to the beach. It is believed that the only relevant conditions are whether or not it will rain and whether or not evening trains will run. Utilities this time are:

|  | Don't go | Go |
|--|----------|-----|
| (rain, train) | 3/4 | 1/2 |
| (rain, no train) | 7/8 | 0 |
| (no rain, train) | 1/8 | 1 |
| (no rain, no train) | 1/2 | 5/8 |

Conditional independence relations that would permit calculation of a unique (maximum entropy) joint distribution over all of the attributes mentioned in the tables, and from which (by marginalization) a unique probability distribution over the four joint conditions for the decision problem could be calculated, are not assumed. Nonetheless, it can be determined from the database that exactly one of the actions is E-admissible. There are infinitely many distributions over the Cartesian product of the domains of the attributes in the database whose marginals coincide with the distributions in the database. Each of these is a solution to a system of 20 linear equations in 48 unknowns. (The solution set is referred to as the *extension* of the database.) The probabilities of any of the 4 rain/train conditions is the sum of 12 of the 48 unknowns. Thus, E-admissibility can be determined as in the previous examples.

We have seen that, for various systems of probabilistic logic, it is not necessary to take into account all of the available sentences (even those that are relevant in the sense of having an effect on the entailed probabilities of the conditions) in order to solve decision problems. Similarly, working with an entire database may introduce unnecessary expense. Anytime algorithms can be devised as well for decision making with probabilistic databases.

The structure of a database, i.e., the set of sets of attributes on which it is defined, is referred to as its *scheme*. The scheme for the database above is

    {{Rain,No Phones},{No Phones,Trains},
        {No Phones,Temp.},{Temp.,Humidity}}.

Scheme $S$ is a refinement of scheme $S'$ iff for each $V \in S$ there exists a $V' \in S'$ such that $V \subseteq V'$. A database may be *projected* onto any scheme that is a refinement of its own. The result is a database whose elements are marginals of its own elements. For example, the projection of the database above onto the scheme {{Trains},{Temperature}} is:

| Trains | $p^5$ | Temperature | $p^6$ |
|--------|-------|-------------|-------|
| yes | 0.5 | high | 0.7 |
| no | 0.5 | med | 0.2 |
|  |  | low | 0.1 |

If $S$ is a refinement of $S'$, then the extension (to any number of attributes) of the projection of a database onto $S'$ is a subset of the extension of the projection onto $S$ [Pittarelli, 1994]. Thus, if an action is E-admissible relative to the set of probabilities over the conditions that can be calculated from a database, then it is E-admissible relative to the probabilities calculated from any projection of the database.

Equivalently, if an action can be determined not to be E-admissible relative to a projection, it can be inferred that it is not E-admissible relative to the original database.

Since the set of E-admissible actions decreases monotonically as schemes become less refined, anytime decision methods are possible for problems in which the set of conditions is the Cartesian product of attribute domains from the database (or can be constructed from the tuples in such a product). Let $V_C$ denote this set of attributes and let $S$ denote the scheme for the database. For purposes of illustration only, a particularly simple-minded approach would be the following: Project first onto

$$\{\{v\}|v \in V_C\}.$$

Next, if necessary, project onto

$$\{V \cap V_C \mid V \in S, \ V \cap V_C \neq \varnothing\}.$$

Next, try

$$\{V \mid V \in S, \ V \cap V_C \neq \varnothing\}.$$

Extend the entire database (or extend its projection onto some scheme that can be identified, at some cost, as producing the same result less expensively [Pittarelli, 1993]) only as a last resort.

For the beach decision problem, $V_C = \{\text{Rain}, \text{Trains}\}$. The projection onto $\{\{v\}|v \in V_C\}$ is

| Trains | $p^5$ | Rain | $p^7$ |
|--------|-------|------|-------|
| yes    | 0.5   | yes  | 0.5   |
| no     | 0.5   | no   | 0.5   |

Both actions are E-admissible relative to the set of joint probabilities compatible with this database.

For this problem,

$$\{V \cap V_C \mid V \in S, \ V \cap V_C \neq \varnothing\} = \{\{v\}|v \in V_C\}.$$

The projection onto $\{V \mid V \in S, \ V \cap V_C \neq \varnothing\}$ is the set of distributions $\{p^1, p^2\}$, above. "Don't go" is identified from this set of distributions as uniquely E-admissible. There exist feasible solutions to the system of inequalities below, but not to the corresponding system for "Go".

$p(Rain = yes, No\ Phones = true, Trains = yes)$
$\qquad + \ p(yes, true, no) = 0.4$
$p(yes, false, yes) \ + \ p(yes, false, no) = 0.1$
$p(no, true, yes) \ + \ p(no, true, no) = 0.2$
$p(no, false, yes) \ + \ p(no, false, no) = 0.3$
$p(yes, true, yes) \ + \ p(no, true, yes) = 0.25$
$p(yes, true, no) \ + \ p(no, true, no) = 0.35$
$p(yes, false, yes) \ + \ p(no, false, yes) = 0.25$
$p(yes, false, no) \ + \ p(no, false, no) = 0.15$
$(\frac{3}{4} - \frac{1}{2}) \times (p(yes, true, yes) + p(yes, false, yes))$
$\qquad + \ (\frac{7}{8} - 0) \times (p(yes, true, no) + p(yes, false, no)) \ +$

$(\frac{1}{8} - 1) \times (p(no, true, yes) + p(no, false, yes))$
$\qquad + \ (\frac{1}{2} - \frac{5}{8}) \times (p(no, true, no) + p(no, false, no)) \ \geq 0.$

Note that even relative to the (projection of) the extension of the entire database there may be more than one E-admissible action. If this is so, and the database contains probability intervals, then Loui's methods [1986] may be applied to narrow them. Alternatives applicable to point-valued probabilistic databases are the variable and structural refinements discussed by Poh and Horvitz [1993] and "coarsenings" of the database scheme. The latter, which includes structural refinement as a special case (i.e. may, but needn't, introduce new variables) requires the assessment of joint probabilities over supersets of the sets of variables contained in the original database scheme. If the old database is a projection of the new database, then the new set of E-admissible actions is a subset of the old.

# 7  CONCLUSION

Anytime decision methods may be devised for use with probabilistic databases, Frisch and Haddawy's anytime deduction system, and Nilsson's probabilistic logic. Common to each of these methods is the generation of a system of linear inequalities the unknowns of which are probabilities of the conditions for a decision problem. Levi's E-admissibility criterion may be applied to the solution set of the system of inequalities. The size of the system of inequalities increases, and the set of admissible actions shrinks, as more of the knowledge base or database is taken into account.

Specific measures of the quality of a decision are not explored. It seems that, for a fixed set of actions under consideration, reasonable measures will be such that the quality of the decision based on a set $E$ of admissible actions will be higher (ignoring the cost of computation) than that of any decision based on a superset of $E$. For each of the methods discussed, actions are eliminated from consideration as computation proceeds. Thus, the quality of a decision (made by choosing an action from the currently admissible set using some criterion other than E-admissibility) increases with time.

Determining which sentences, or projections of a database, will eliminate the greatest number of actions at the least cost, and whether it is worth the effort to consider additional sentences or projections at all, is a difficult problem which remains for future research.

### References

[Barbara et al., 1993] D. Barbara, H. Garcia-Molina, and D. Porter. The management of probabilistic data. *IEEE Trans. on Knowledge and Data Engineering, v. 4*, pp. 387-402.

[Boddy and Dean, 1989] M. Boddy and T. Dean. Solving time-dependent planning problems. *Proc. IJCAI-89*, Morgan Kaufmann, pp. 979-984.

[Cavallo and Pittarelli, 1987] R. Cavallo and M. Pittarelli. The theory of probabilistic databases, *Proc. 13th Conf. on Very Large Databases*, Morgan Kaufmann, pp. 71-81.

[Deutsch-McLeish, 1990] M. Deutsch-McLeish. An investigation of the general solution to entailment in probabilistic logic. *Int. J. of Intelligent Systems, v. 5*, pp. 477-486.

[Elster, 1989] J. Elster. *Solomonic Judgements*. Cambridge University Press.

[Frisch and Haddawy, 1992] A. Frisch and P. Haddawy. Anytime deduction for probabilistic logic. Technical Report UIUC-BI-AI-92-01, Beckman Institute, Univ. of Illinois, Urbana. To appear in *Artificial Intelligence*.

[Grosof, 1986] B. Grosof. An inequality paradigm for probabilistic knowledge. In L. Kanal and J. Lemmer, Eds., *Uncertainty in Artificial Intelligence*, North-Holland.

[Horvitz et al., 1989] E. Horvitz, G. Cooper and D. Heckerman. Reflection and action under scarce resources: theoretical principles and empirical study. *Proc. IJCAI-89*, Morgan Kaufmann, pp. 1121-1127.

[Jaynes, 1982] E. T. Jaynes. The rationale of maximum-entropy methods. *Proc. of the IEEE, v. 70*, pp. 939-952.

[Jeffrey, 1976] R. Jeffrey. Savage's omelet. In F. Suppe and P. Asquith, Eds., *PSA 1976, v. 2*, Philosophy of Science Association.

[Kane, 1990] T. Kane. Enhancing the inference mechanism of Nilsson's probabilistic logic. *Int. J. of Intelligent Systems, v. 5*, pp. 487-504.

[Kane, 1991] T. Kane. Reasoning with maximum entropy in expert systems. In W. T. Grandy, Jr. and L. Schick, Eds., *Maximum Entropy and Bayesian Methods*, Kluwer.

[Kyburg, 1992] H. E. Kyburg, Jr. Getting fancy with probability. *Synthese, v. 90*, pp. 189-203.

[Levi, 1980] I. Levi. *The Enterprise of Knowledge*. MIT Press.

[Loui, 1986] R. Loui. Decisions with indeterminate probabilities. *Theory and Decision, v. 21*, pp. 283-309.

[MacQueen and Marschak, 1975] J. MacQueen and J. Marschak. Partial knowledge, entropy, and estimation. *Proc. Nat. Acad. Sci., v. 72*, pp. 3819-3824.

[Nilsson, 1986] N. Nilsson. Probabilistic logic. *Artificial Intelligence, v. 28*, pp. 71-87.

[Piepel, 1983] F. Piepel. Calculating centroids in constrained mixture experiments. *Technometrics, v. 25*, pp. 279-283.

[Pittarelli, 1991] M. Pittarelli. Decisions with probabilities over finite product spaces. *IEEE Trans. SMC, v. 21*, pp. 1238-1242.

[Pittarelli, 1993] M. Pittarelli. Probabilistic databases and decision problems: results and a conjecture. *Kybernetika, v. 29*, pp. 149-165.

[Pittarelli, 1994] M. Pittarelli. An algebra for probabilistic databases. *IEEE Trans. on Knowledge and Data Engineering, v. 6*, pp. 293-303.

[Poh and Horvitz, 1993] K. L. Poh and E. Horvitz. Reasoning about the value of decision-model refinement: methods and application. *Proc. of the 9th Conf. on Uncertainty in Artificial Intelligence*, pp. 174-182.

[Quinlan, 1983] R. Quinlan. Inferno: a cautious approach to uncertain inference. *The Computer Journal*, v. 26, pp. 255-269.

[Snow, 1991] P. Snow. Compressed constraints in probabilistic logic and their revision. *Proc. 7th Conf. on Uncertainty in Artificial Intelligence*, Morgan Kaufmann, pp. 386-391.

[Starr, 1966] M. Starr. A discussion of some normative criteria for decision-making under uncertainty. *Industrial Management Review, v. 8*, pp. 71-78.